\documentclass[sigconf]{acmart}

\usepackage{subfig}
\usepackage{multirow, booktabs}
\usepackage{siunitx}
\usepackage{amsmath}
\usepackage{pifont}

\AtBeginDocument{%
  \providecommand\BibTeX{{%
    \normalfont B\kern-0.5em{\scshape i\kern-0.25em b}\kern-0.8em\TeX}}}

\copyrightyear{2022}
\acmYear{2022}
\setcopyright{acmlicensed}\acmConference[GECCO '22]{Genetic and Evolutionary Computation Conference}{July 9--13, 2022}{Boston, MA, USA}
\acmBooktitle{Genetic and Evolutionary Computation Conference (GECCO '22), July 9--13, 2022, Boston, MA, USA}
\acmPrice{15.00}
\acmDOI{10.1145/3512290.3528744}
\acmISBN{978-1-4503-9237-2/22/07}

\begin{document}

\title{RankNEAT: Outperforming Stochastic Gradient Search in Preference Learning Tasks}

\author{Kosmas Pinitas}
\affiliation{%
\institution{University of Malta}
\city{Msida}
\country{Malta}}
\email{kosmas.pinitas@um.edu.mt}

\author{Konstantinos Makantasis}
\affiliation{%
\institution{University of Malta}
\city{Msida}
\country{Malta}}
\email{konstantinos.makantasis@um.edu.mt}

\author{Antonios Liapis}
\affiliation{%
\institution{University of Malta}
\city{Msida}
\country{Malta}}
\email{antonios.liapis@um.edu.mt}

\author{Georgios N. Yannakakis}
\affiliation{%
\institution{University of Malta}
\city{Msida}
\country{Malta}}
\email{georgios.yannakakis@um.edu.mt}

\renewcommand{\shortauthors}{Pinitas et al.}

\begin{abstract}
Stochastic gradient descent (SGD) is a premium optimization method for training neural networks, especially for learning objectively defined labels such as image objects and events. When a neural network is instead faced with subjectively defined labels---such as human demonstrations or annotations---SGD may struggle to explore the deceptive and noisy loss landscapes caused by the inherent bias and subjectivity of humans. While neural networks are often trained via preference learning algorithms in an effort to eliminate such data noise, the \emph{de facto} training methods rely on gradient descent. Motivated by the lack of empirical studies on the impact of evolutionary search to the training of preference learners, we introduce the RankNEAT algorithm which learns to rank through neuroevolution of augmenting topologies. We test the hypothesis that RankNEAT outperforms traditional gradient-based preference learning within the affective computing domain, in particular predicting annotated player arousal from the game footage of three dissimilar games. RankNEAT yields superior performances compared to the gradient-based preference learner (RankNet) in the majority of experiments since its architecture optimization capacity acts as an efficient feature selection mechanism, thereby, eliminating overfitting. Results suggest that RankNEAT is a viable and highly efficient evolutionary alternative to preference learning.
\end{abstract}





\maketitle

\section{Introduction}\label{sec:introduction}

Forms of gradient descent are the natural choice of optimization method for training deep neural networks to predict objectively defined labels in tasks such as image and speech recognition, fraud detection, and event prediction. Over the last few years, we have witnessed a rapidly growing interest in the use of neural networks that are able to classify subjectively defined labels. This family of learning-to-rank or preference learning algorithms \cite{furnkranz2011preference} that train neural networks---such as RankNet \cite{burges2005learning}, DeepRank \cite{pang2017deeprank} and LambdaMART \cite{burges2010ranknet}---yield good performance by relying primarily on gradient descent methods. Subjectively defined labels, however, including human demonstrations (e.g. creative tasks, navigation traces and paths) or human annotations (e.g. of emotion or aesthetics) yield highly complex, deceptive and noisy loss landscapes for a neural network to learn. Assuming that the plasticity of neuroevolutionary processes would be beneficial for such loss landscapes, in this paper we test the hypothesis that evolutionary search would be a better optimizer for neural network training in preference learning (PL) tasks compared to stochastic gradient descent (SGD).

To test our hypothesis, this paper explores the efficacy of neuroevolutionary search in PL tasks by building on the efficient and popular RankNet \cite{burges2005learning} architecture and enhancing its search capacity through neuroevolution. In particular, we introduce a novel algorithm named \emph{RankNEAT} that relies on the Siamese neural network architecture of RankNet and learns to rank via NeuroEvolution of Augmenting Topologies (NEAT) \cite{stanley2002evolving}. Unlike traditional gradient-based PL methods, RankNEAT resembles the process of plasticity \cite{fair2008brain}, which induces changes in both the coupling strength and the spatial organization of synapses in biological neural networks. RankNEAT learns to rank subjectively defined labels with high degrees of accuracy through its ability to optimize the synaptic parameters such as the network's weights and the edge architecture simultaneously. We test RankNEAT (neuroevolution) and compare it against the vanilla RankNet (stochastic gradient decent) in the task of player affect modeling across three games, using the AGAIN \cite{melhart2021affect} dataset of arousal-annotated gameplay videos. Player modeling \cite{yannakakis2018artificial} is an important subfield in game research since it promotes the development of reliable human computer interaction systems and consequently improves the users' experience. Our current approach feeds images of gameplay to a pretrained vision transformer, while the last fully-connected layer of the network is then trained to predict ordinal values of arousal, using RankNet or RankNEAT. Results indicate that RankNEAT is superior to SGD (RankNet) in training PL models of arousal in the majority of experiments performed. Our key findings suggest that RankNEAT is a viable PL paradigm which achieves comparable or significantly higher performances to RankNet. In this first experiment, RankNEAT optimizes the edge topology of the networks' last layer, resembling an evolutionary feature selection strategy that eliminates unnecessary features from the observed input space. Additional studies should explore how RankNEAT performs in other subjectively defined tasks and hyperparameter setups, such as increasing the topological complexity.

This paper is novel in many ways. First, to the best of our knowledge, this is the first time a NEAT-based preference learner is introduced, combining a traditional learning-to-rank neural network architecture with neuroevolution. Second, RankNEAT is tested broadly across three dissimilar games from the same genre showcasing the robustness of the method for affect modeling. Third, the proposed approach is compared thoroughly against SGD (RankNet) across different games and hyperparameters. Finally, RankNEAT is combined with vision transformers (pretrained on ImageNet) enabling us to offer general-purpose representations for solving tasks with subjectively defined labels.




\section{Related Work}\label{sec:relatedwork}

This section surveys related work on the performance comparison of evolutionary algorithms and gradient descent for training neural networks (see Section \ref{s:2.1}) and on the intersection of evolutionary search and affect modeling (see Section \ref{s:2.2}). 

\subsection{Evolution versus Backpropagation for Neural Network Training}\label{s:2.1}

Although SGD is currently the most widely applied training algorithm for neural networks, there has been a rapidly growing interest in employing evolutionary algorithms for optimizing deep learning models over the last years \cite{zhao1996evolutionary,yao1997new,stanley2002evolving,shinozaki2015structure}. 
Evolution and gradient descent through backpropagation (BP) are, however, fundamentally different and thus their comparison is a challenging task that numerous studies have tried to tackle. Indicatively, Mandischer \cite{mandischer2002comparison} pitted evolutionary strategies (ES) against BP for neural network training on several benchmark problems, evaluating them based on the computational effort required to reach a certain error limit and their ability to converge. Results showed that while ES were good for training neural networks with non-differentiable activation functions, they still cannot compete with BP in large-scale problems. Siddique et al. \cite{siddique2001training} proposed a genetic algorithm (GA) capable of outperforming BP in function approximation in terms of convergence. Sexton et al. \cite{sexton1998toward} compared a GA and BP on in-sample, interpolation, and extrapolation data in terms of root-mean-square error, number of epochs, and execution time. Results showed that GAs can be employed to strike a balance between model over-parameterization and model robustness.
Gupta et al. \cite{gupta1999comparing} compared GAs and BP in terms of effectiveness, ease-of-use, and efficiency for training neural networks, showing that the former can provide better results in a chaotic time series problem. Gudise et al. \cite{gudise2003comparison} conducted a comparative study which demonstrated that the weights of a feedforward neural network tend to converge faster with the particle swarm optimization than with BP when it comes to function approximation. Finally, Sexton et al. \cite{sexton2000reliable} compared evolution and BP across ten real-world classification problems, showing that BP reached a higher classification error on average.
Yannakakis et al. \cite{yannakakis2007emerging} employed supervised and genetic approaches to study the emergence of cooperative behavior among agents in a complex simulated environment and demonstrated that a genetic approach based on rewarding and minimal communication resulted in more efficient computational models of multi-agent spatial organization than supervised learning mechanisms. Zhang et al. \cite{zhang2017relationship} conducted several MNIST-based experiments in order to shed light on the relationship between the OpenAI ES and SGD by measuring the correlation between the approximated gradients computed by the algorithms and developing an SGD-based proxy for ES. The results obtained by the ES proxy are identical with those obtained by ES, and consequently, it holds that SGD with noise is equivalent to ES. Morse \& Stanley \cite{morse2016simple} compared an evolutionary algorithm that evaluates individuals on a small number of training samples per generation and SGD on several benchmarks and showed that the former could optimize large neural networks about as fast and effectively as the latter. 

In this work, we extend the literature by comparing neuroevolution and SGD performance in preference learning problems for the first time. In particular, we introduce a NEAT-based preference learner capable of predicting player arousal from gameplay footage and compare its performance with that obtained via SGD. Our results show that combining the topological and global optimization properties of NEAT with the Siamese network architecture of the traditional RankNet can result in robust learning-to-rank models that outperform the BP models trained via SGD.


\subsection{Modeling Affect via Evolutionary Search}\label{s:2.2}

Affective computing is the study of emotions, their manifestations and expressions, and the ways to capture (model) them computationally \cite{picard2000affective}. 
While research at the intersection of affective computing and evolutionary algorithms has been active over the last decade, studies in the literature are still relatively sparse. For instance, Martinez et al. \cite{martinez2010genetic} presented a genetic search-based feature selection method for improving the accuracy of the affective models, comparing it against sequential forward feature selection and random search in a game survey dataset. Tahir et al. \cite{tahir2020novel} introduced a binary chaotic genetic algorithm for feature selection, which achieved scores two times higher than a baseline genetic algorithm in identifying seven emotional states. Finally, Alvarez et al. \cite{alvarez2006feature} employed artificial evolution to select speech feature subsets that optimize the success rate of emotion recognition.

When it comes to games, the domain we study in this paper, player modeling \cite{yannakakis2018artificial} refers to the study of models that accurately predict how a player behaves and feels while playing a game. 
Affect models based on gameplay can provide valuable insights into how players interact with games. As input of such models, most studies apply domain knowledge by manually authoring high-level hand-crafted gameplay features. For instance, Frommel et al. \cite{frommel2018towards} employed the input parameters on a graphics tablet and in-game performance to detect the players’ current emotional state. Similarly Melhart et al. \cite{melhart2021towards} showed that hand crafting features that describe the player’s input, the artificial agents’ actions, and the gameplay context on a high level can yield general models of player arousal.

Although domain knowledge may lead to remarkable results, hand-crafted features do not necessarily reduce the data needs of the algorithms but do introduce a critical data preprocessing step. Automated feature extraction, on the other hand, may address such issues. Literature in this vein is fairly sparse. Ng et al. \cite{ng2015deep} used a deep Convolutional Neural Network (CNN) pretrained on the generic ImageNet dataset to perform emotion recognition on small datasets. Makantasis et al. \cite{makantasis2019pixels} employed three CNN architectures to predict player arousal from gameplay footage, showcasing that a mapping between gameplay video streams and the player's arousal exists. The same authors also introduced a methodology for predicting arousal from audiovisual features and demonstrated that fusing high-level pixel and audio representations can yield highly accurate models of affect \cite{makantasis2021pixels}. Finally, the study of Martinez et al. \cite{martinez2013learning} seems to be the first to introduce a deep PL methodology for predicting emotional states from physiological signals. In particular, they showed that using auto-encoders and CNNs to find a mapping from raw signals to learnable features can outperform ad-hoc feature extraction and selection. 

In contrast to all aforementioned studies, in this work we employ a pretrained Vision Transformer to extract high-level representations from gameplay footage and fine-tune our RankNEAT model to construct player arousal models for three platformer games. We also compare the behavior of evolutionary PL against gradient-based PL. Results verify that RankNEAT outperforms RankNet in most experiments performed due to the global optimization capabilities of the former. At the same time, its architecture search capacity corresponds to an effective mechanism of feature elimination.

\begin{figure}[!tb]
\centering
\includegraphics[width=\linewidth]{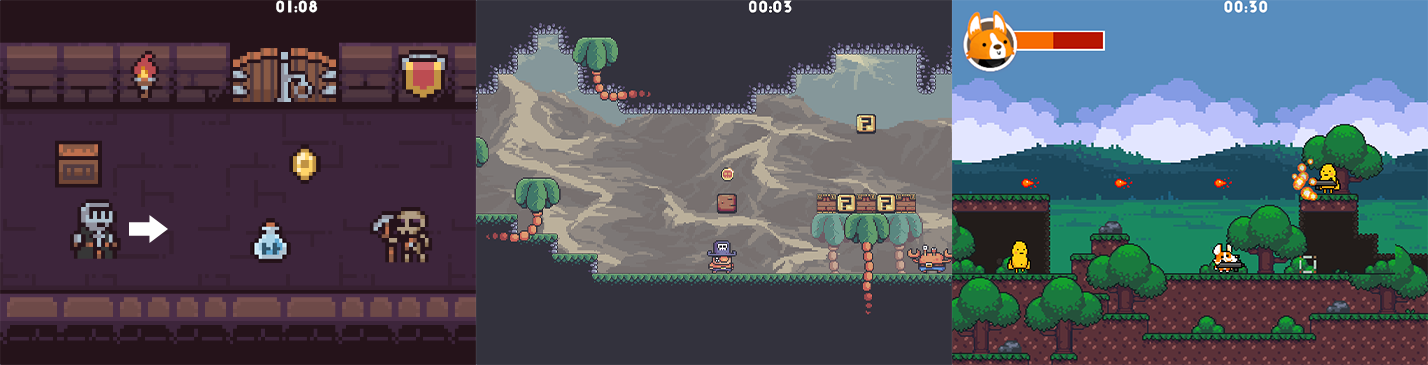}
\caption{The three games used in this study. From left to right: Endless, Pirates!, Run'N'Gun.}
\label{fig:games}
\end{figure}

\section{Case Study: Predicting Player Affect}

The neuroevolutionary learn-to-rank methodology proposed in this paper is tested on a challenging dataset of three games which includes many players' gameplay and emotion annotations. The games are created for the purposes of general affect modeling and are part of the AGAIN dataset \cite{melhart2021affect}. In this paper we focus on the three games of the platformer genre featured in AGAIN as they offer sufficiently diverse gameplay properties without needing excessive computation for experimental validation. The three games are shown in Fig. \ref{fig:games} and include \textbf{Endless}, an infinite runner where players must avoid obstacles while automatically moving ever rightward, \textbf{Pirates!}, a jumping platformer similar to \emph{Super Mario Bros} (Nintendo, 1985), and \textbf{Run'N'Gun}, a more complex game which requires players to move while aiming and shooting at enemies. All games have arcade-style controls of varying complexity (with Run'N'Gun being the most complex), assign a score to the player for in-game actions, and finish after two minutes for the purposes of data collection. 

\begin{figure}[!t]
\centering
\includegraphics[width=1.0\linewidth]{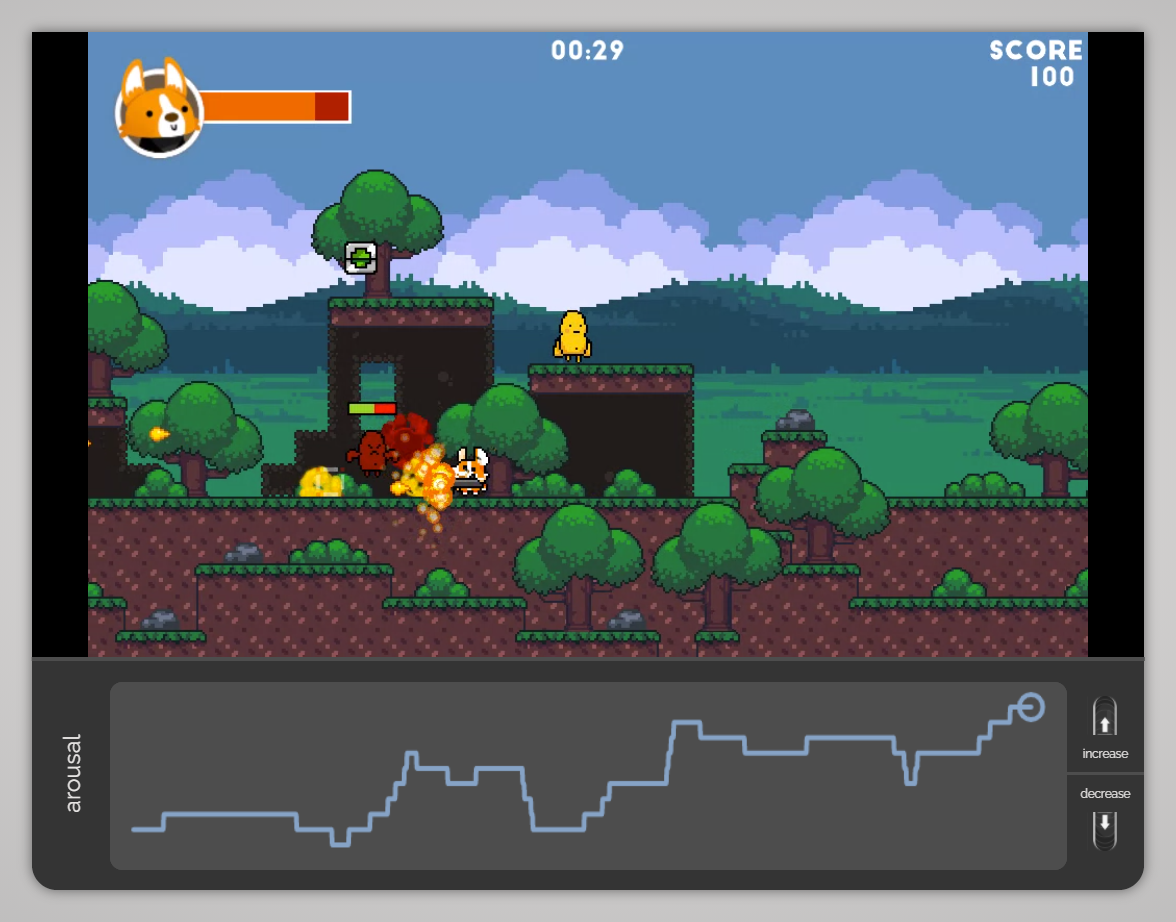}
\caption{Arousal annotation of a player's gameplay footage using the time-continuous unbounded RankTrace protocol.}
\label{fig:pagan}
\end{figure}

The AGAIN dataset was collected through Mechanical Turk, with players first playing and then annotating each game. Annotation was done using a stimulated recall protocol, showing the player's own gameplay as a recorded video and requiring them to provide moment-to-moment annotation of arousal using the RankTrace annotation tool \cite{lopes2017ranktrace}. Figure \ref{fig:pagan} shows the arousal annotation trace: the player can keep increasing or decreasing their arousal annotation (unbounded) and can view their entire annotation so far. 

The arousal annotations are preprocessed before being used for modelling tasks. First, the trace is normalized to $[0,1]$ with min-max normalization. Due to the reaction time between a stimulus and a player's emotional response, we processed the data into time windows of 3 seconds. In particular, we calculate the mean arousal value of 3-second time window which we then use as the subjectively defined label for training (see Section \ref{sec:preference_learning}). Gameplay videos are captured at 24Hz, resulting in 72 frames per 3-second window. Each gameplay frame is rescaled to a 224 by 224 pixel RGB image, and the 72 frames of the 3-second window are processed iteratively through a Vision Transformer (see Section \ref{sec:vim}).

When aligning arousal time windows and gameplay frames' time windows, we apply 1 second lag (shifting the annotations 1 second back compared to the video data) to simulate delays in the annotation process \cite{mariooryad2013analysis,melhart2021towards}. After processing the data into 3-second windows and data cleanup (e.g. videos with missing frames due to errors during video recording), the dataset across all three games consists of 262 gameplay videos corresponding to 111 different players. In particular, 103 gameplay videos remain for Endless ($4,120$ time windows), 92 videos for Pirates! ($3,680$ time windows), and 67 videos for Run'N'Gun ($2,680$ time windows). Each video within the same game corresponds to a different player, which is important for cross-validation purposes (see Section \ref{sec:results}).

\begin{figure}[!tb]
\centering\includegraphics[width=1.0\linewidth]{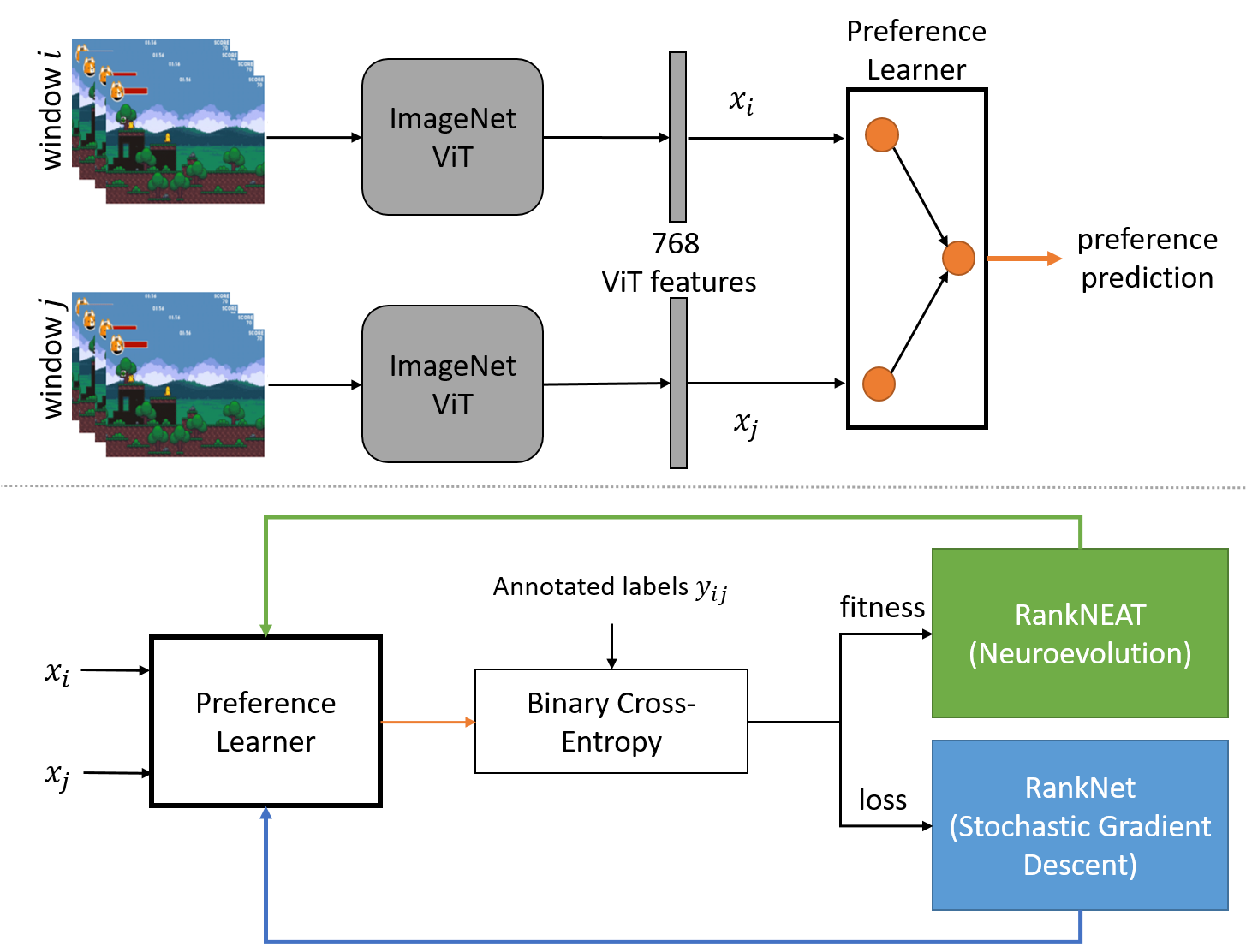}
\caption{Illustration of the preference learning architecture (top image) and training methods (bottom image) used. ImageNet ViT (see Section \ref{sec:vim}) is not trainable; only the preference learner is trained. Gameplay footage feeds the Siamese PL architecture and the binary cross-entropy is calculated on the difference between the network's output and respective preference (annotated) label $y_{ij}$ (see Section \ref{sec:preference_learning}). 
The training parameters of the preference learner are optimized either: (a) via RankNEAT (Section \ref{sec:preference_learning}) where the binary cross-entropy becomes the fitness, or (b) via RankNet \cite{burges2005learning} where the binary cross-entropy serves traditionally as the loss function.}
\label{fig:methodology}
\end{figure} 

\section{Methodology}\label{sec:methodology}

This section describes the main components of the algorithms examined in this paper including the Vision Transformers \cite{dosovitskiy2020image} used to extract high-level features from the video data and the two training methods used for our preference learning task: SGD and neuroevolution. An overview of our approach is presented in Fig. \ref{fig:methodology}.

\subsection{Vision Transformer}\label{sec:vim}

A Transformer is an architecture that utilizes an attention mechanism to discover dependencies between input and output. Although Transformers still employ an encoder and decoder, they eschew recurrence and thus require less training time while achieving better results than other sequence transduction models. The Vision Transformer (ViT) is a Transformer-based architecture for image classification tasks, using a single image as input and mapping it to a high-level vector representation, which, in turn, is fed to a multi-linear perceptron responsible for conducting the classification task. 

In this study, we use a ViT pre-trained on ImageNet 1K \cite{deng2009imagenet} as a backbone model to retrieve high-level vector representations of gameplay frame sequences. Since each time window consists of 72 frames, we replicated the weights of the first layer of ViT 72 times to account for input mismatch. Through this process, the $72\times224\times224\times3$ tensor of pixel values bound between $[0,1]$ is transformed into a vector of 768 real values that represent higher representations of the gameplay video segment (see Fig. \ref{fig:methodology}).

\subsection{Preference Learner}\label{sec:preference_learning}

Preference learning involves learning to distinguish data points in an ordinal manner \cite{furnkranz2011preference}, and thus can be applied to any supervised problem as long as the labels represent ordinal relationships. Since emotions are ordinal by nature \cite{yannakakis2018ordinal,yannakakis2017ordinal},
in this study we develop a preference learner based on the RankNet architecture \cite{burges2005learning} to predict players' arousal using ViT representations of gameplay footage. Our arousal models are trained on pairs of gameplay windows.

Specifically, we formulate the arousal prediction task as a PL problem in the following way. Let us denote as $\mathcal X$ the space of ViT representations of gameplay footage windows and the data corresponding to the $k$-th gameplay video as $\mathcal D_k = \{(x_i, \lambda_i)\}_{i=1}^N$, 
where $x_i \in \mathcal X$, and $\lambda_i \in [0,1]$ stands for the arousal annotation of the gameplay's $i$-th window. To employ a PL model, we transform $\mathcal D_k$ to $\tilde{\mathcal D}_k = \{(x_i, x_j, y_{ij})\}_{i,j=1}^N$,
where $y_{ij}$ equals 1 if $\lambda_i - \lambda_j > P_t$ and 0 if $\lambda_j - \lambda_i > P_t$.
It should be noted that when $|\lambda_i - \lambda_j| < P_t$ the pair $(x_i, x_j)$ is not included in the dataset $\tilde{\mathcal D}_k$. The preference threshold $P_t$ controls whether or not the difference between the labels qualifies as a preference, while parameter $k$ emphasizes the fact that pairs are produced with datapoints belonging to the same gameplay footage. Finally, the above data transformation procedure results in a perfectly balanced binary classification dataset.

As mentioned above, we adopt the RankNet model \cite{burges2005learning} for addressing the aforementioned PL problem. RankNet employs a neural network that receives as input pairs $(x_i,x_j)$ and their respective labels $y_{ij}$ and outputs $z_{ij}=f(x_i)-f(x_j)$, where $f$ is a scalar function computed by the neural network. RankNet training aims to estimate the parameters of the neural network that minimize the binary cross-entropy loss of $\sigma(z_{ij})$ with respect to $y_{ij}$, where $\sigma(\cdot)$ is the sigmoid logistic function. In our experiments, we consider linear functions $f$, that is neural networks with no hidden layers, and we estimate the parameters of $f$ using two fundamentally different optimization methods: SGD with backpropagation---as traditionally employed in RankNet training---and neuroevolution, as described below through RankNEAT.

\subsubsection*{RankNEAT} 

NeuroEvolution of Augmenting Topologies is an established algorithm \cite{stanley2002evolving} which goes beyond earlier approaches to neuroevolution which represented only the weights of the network as a vector in the genotype. While the typical NEAT algorithm starts from a minimal network (with only input and output nodes) and expands it with new nodes and edges, in this paper we use a simplified version of NEAT which does not add new nodes and thus does not expand the size of the network. It should be noted, however, that other features of NEAT which are crucial to its success, such as speciation and custom operators for adding and removing edges are maintained. In RankNEAT we use NEAT to train our RankNet model by optimizing the parameters of the linear function $f(\cdot)$ via weight mutations, crossover, adding or removing edges. Hence, its behavior resembles a feature elimination mechanism which is essentially the same as setting the weight parameters of the deleted edges to zero.

We use the standard implementation of NEAT-Python \cite{neat-python} for running evolution. The initial population consists of $p$ fully connected RankNet networks with random weights, which are evaluated and then split into species based on their topological similarities. The fitness of each individual is calculated by processing all pairs in the training set through the ViT and RankNet. Each pair consists of two frame sequences $(x_i, x_j)$ and one ground truth preference $(y_{ij})$; each network is processed through the ViT and RankNet to derive $f(x_i)$ and $f(x_j)$ and finally to calculate the negative binary cross-entropy of the produced $\sigma(z_{ij})$. The mean of all cross-entropy scores for each pairing forms the fitness of the network and informs the selection of parents to mate and mutate. 

\section{Results}\label{sec:results}

This paper aims to leverage neuroevolution for preference learning, assuming that its global optimization strategy may prove beneficial compared to gradient descent. Thus, the performance metric in our experiments is the accuracy in predicting the ranking between unseen pairs of gameplay footage windows. Specifically, we use a ten-fold cross-validation strategy for splitting the data into training and test sets. We ensure that data in the test set belongs to players that are absent from the training set. Therefore, we follow a leave-$X$-participants out method for cross-validation, where $X$ is set between 6 and 11 participants depending on the game and fold. To address the randomness of weight initialization, genetic operators, and SGD, results are averaged across 5 independent runs \cite{hastie01statisticallearning} throughout the paper (including the 95\% confidence interval between these 5 runs).

Due to the many hyperparameters of RankNet and RankNEAT, we perform a sensitivity analysis in Section \ref{sec:results_tuning} and report the main findings. Using the best parameters, Section \ref{sec:results_versus} compares the performance of RankNet and RankNEAT for the three games, attempting to provide a fair ground of comparison in terms of computational effort. Throughout the experiments, we perform three tests per game by varying the preference threshold ($P_t$) between $0.15$, $0.25$ and $0.50$. Higher threshold values can be more dependable in terms of the accuracy of the ranking but lead to significantly smaller datasets for training and testing.

\newcommand{\resw}{0.5\linewidth}
\begin{figure}[!tb]

\subfloat[Endless, RankNEAT]{\includegraphics[width=\resw]{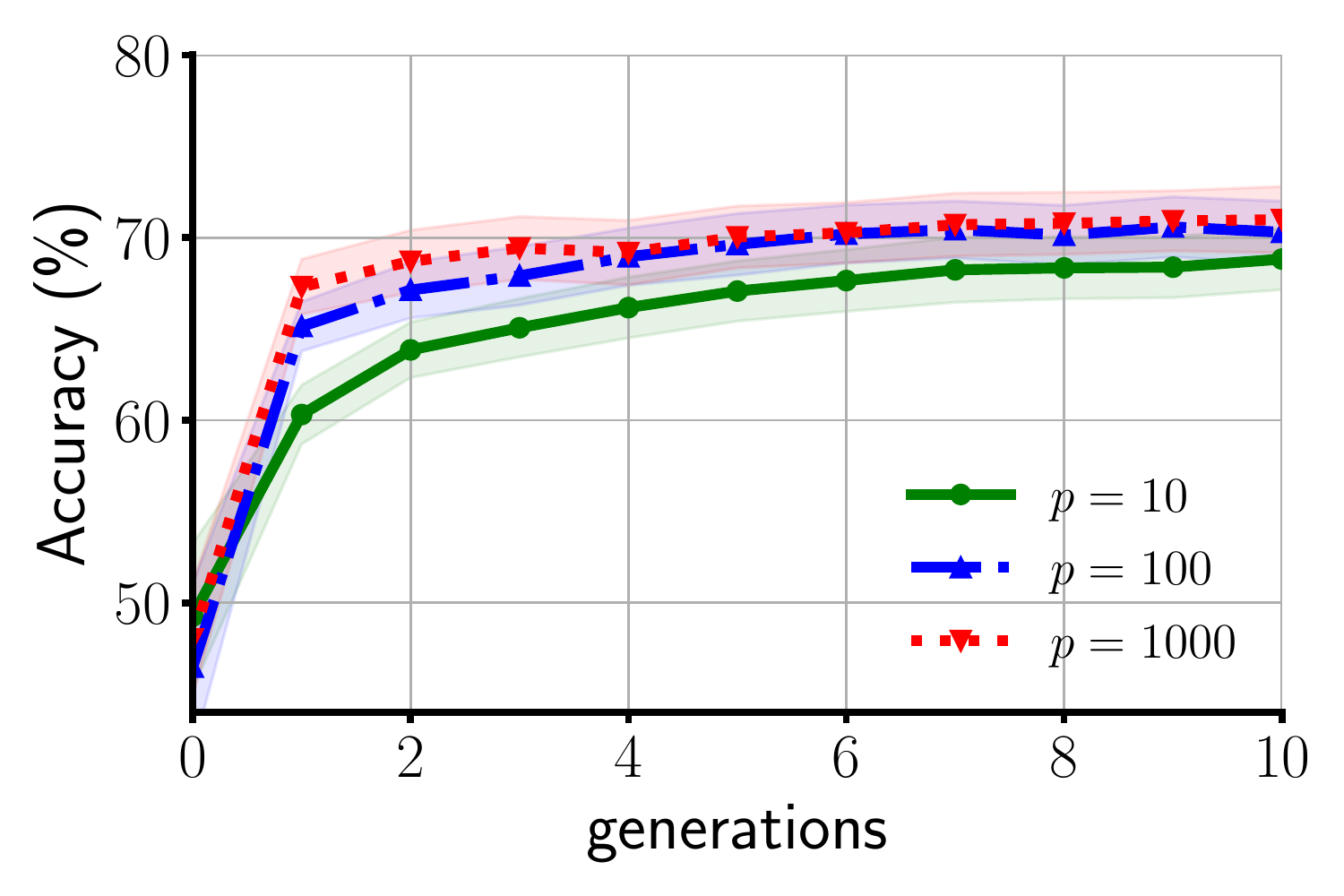}\label{fig:4cA}}
\subfloat[Endless, RankNet]{\includegraphics[width=\resw]{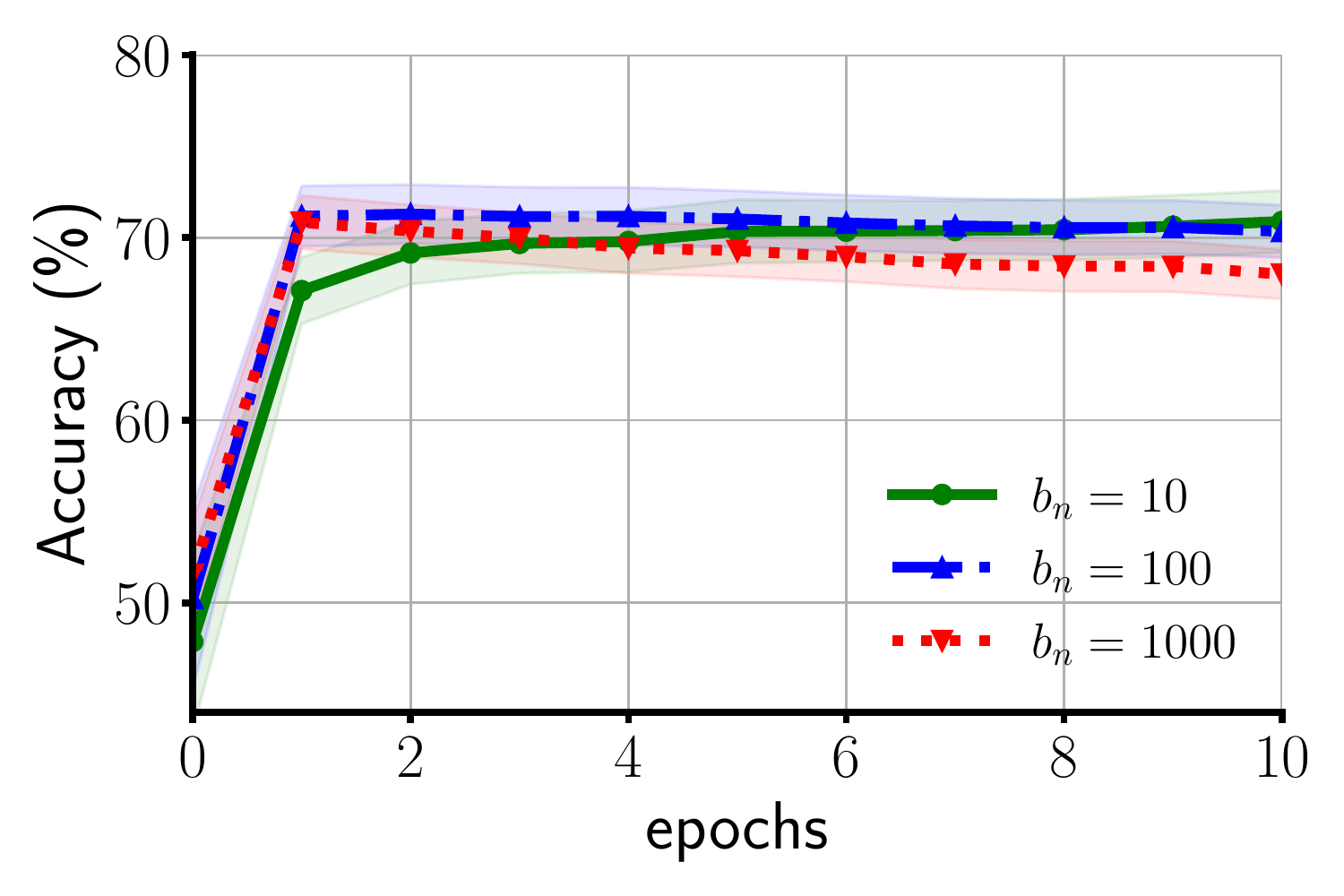}\label{fig:4b}}\\

\subfloat[Pirates!, RankNEAT]{\includegraphics[width=\resw]{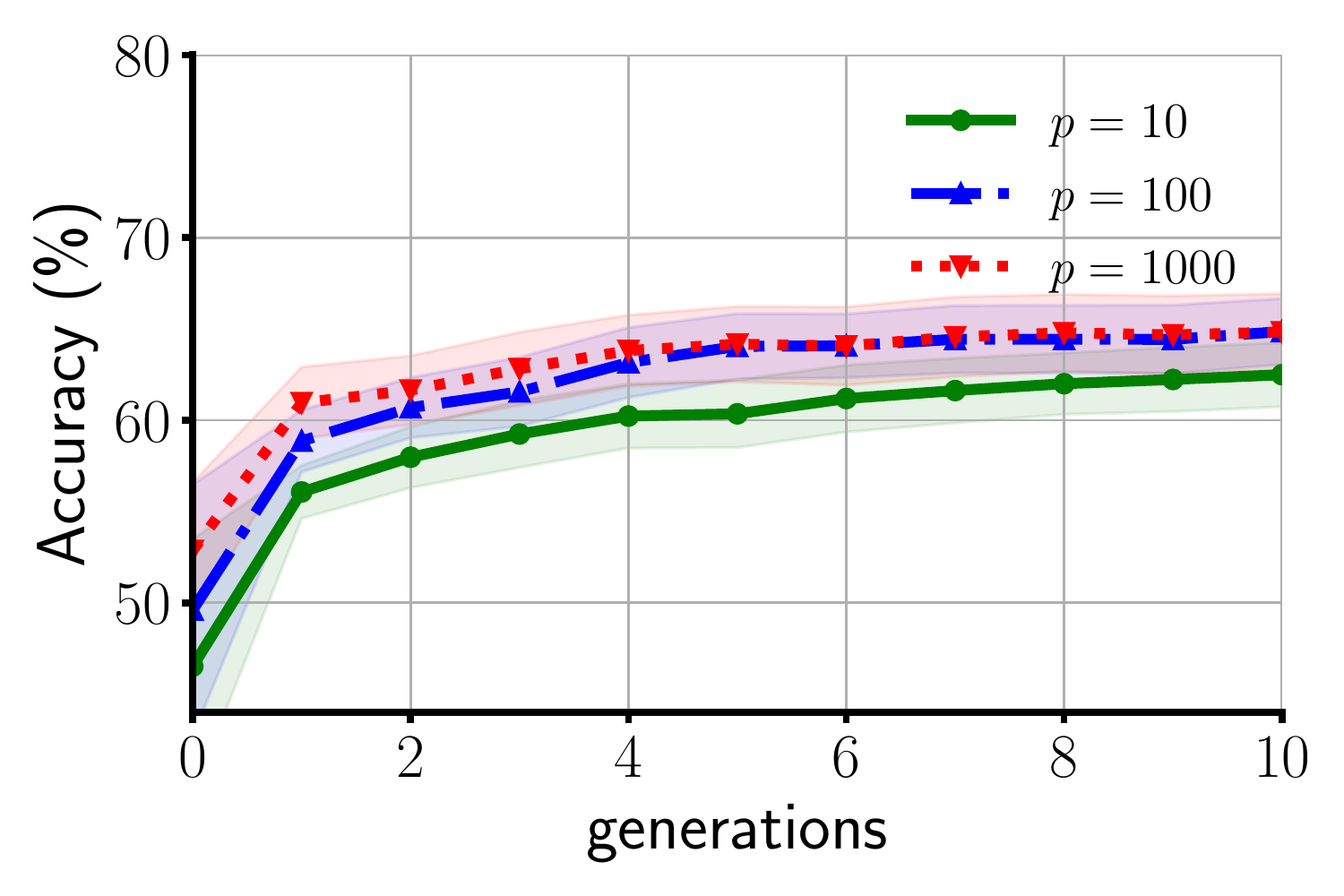}\label{fig:4c}}
\subfloat[Pirates!, RankNet]{\includegraphics[width=\resw]{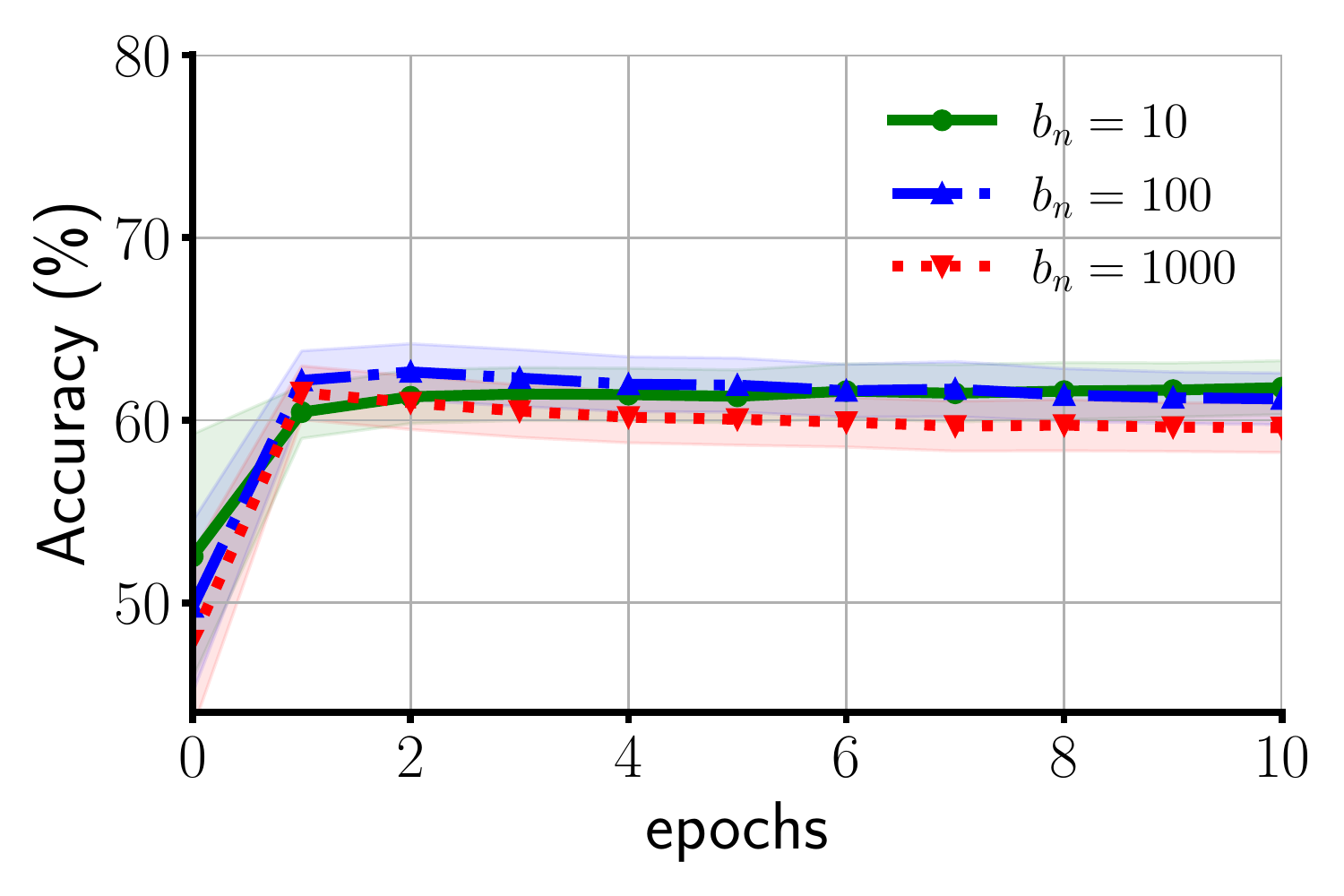}\label{fig:4d}}\\

\subfloat[Run'N'Gun, RankNEAT]{\includegraphics[width=\resw]{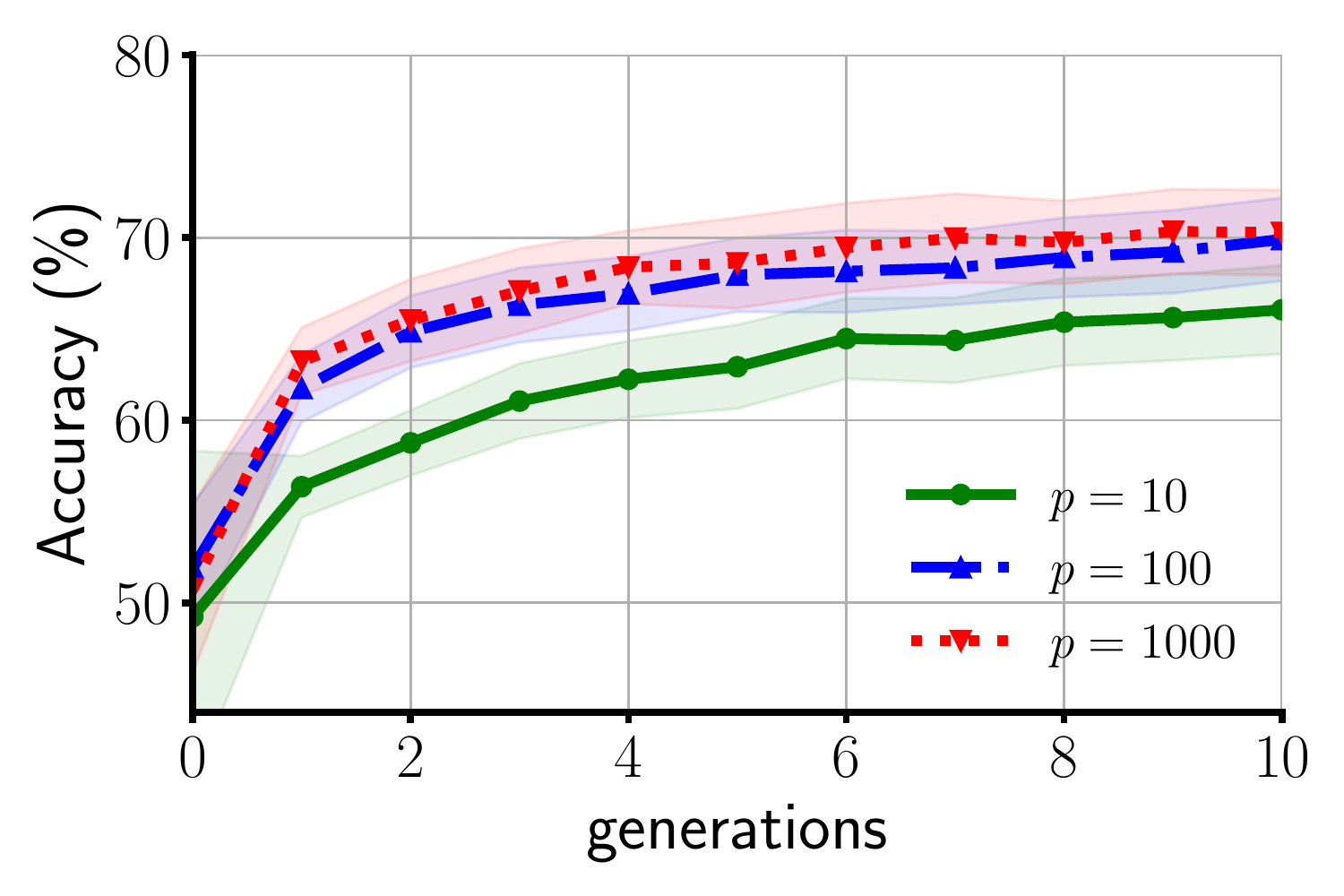}\label{fig:4e}}
\subfloat[Run'N'Gun, RankNet]{\includegraphics[width=\resw]{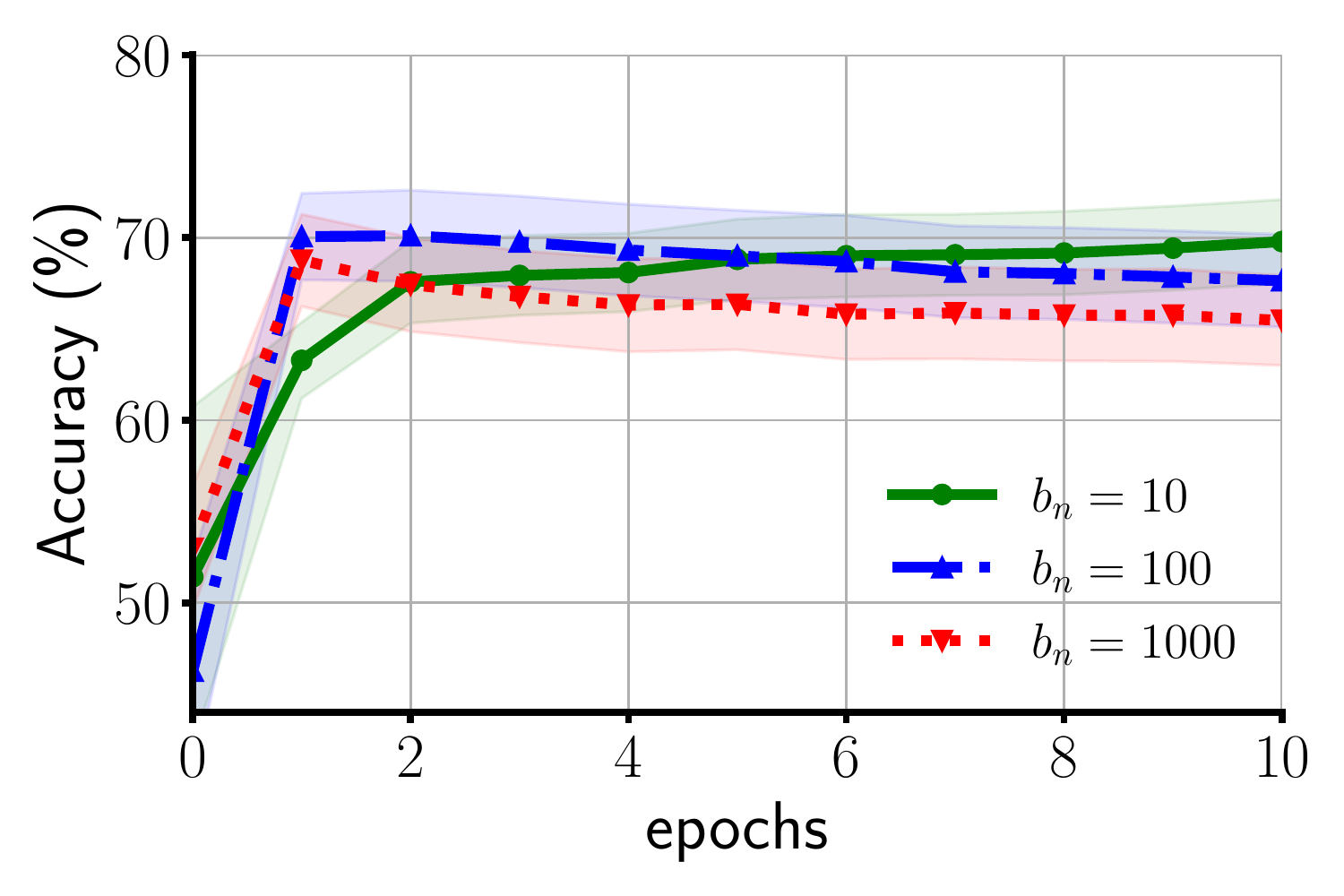}\label{fig:4f}}\\

\caption{Impact of the population and batch size to the performance of the two algorithms.}
\label{fig:acc_param_tuning}
\end{figure}

\subsection{Parameter Tuning}\label{sec:results_tuning}

Several training hyperparameters control the behavior of both NEAT and SGD as optimizers. Parameter setting is often achieved through empirical trial-and-error processes. In terms of RankNet, we tune the batch size since the benefits of the adjustment of this parameter is two-fold. On the one end, the batch size is inversely proportional to the number of updates per epoch, affecting the speed of the training process. On the other end, the ratio of learning rate to batch size is a key element influencing the SGD dynamics \cite{jastrzkebski2017three}.
When it comes to RankNEAT, there is no single correct choice of parameters for all problems due to interdependencies between hyperparameters such as population size and crossover \cite{de1990analysis}. Although the compatibility threshold ($c_t=3$), elitism per species ($e_{ps}=2$), and mutation rates ($0,0.5$ for nodes and edges, respectively) were tuned according to some preliminary experiments, the population size $p$ was adjusted based on a more systematic approach since it influences both the training time and the robustness of the learner \cite{rylander2002optimal}. This section details our experiments on the three game test-beds for determining the optimal population size $p$ and batch number $b_n$. It should be noted that other hyperparameters such as the learning rate for SGD, the compatibility coefficients and the survival threshold for NEAT were kept at their default values from their respective libraries. For space considerations we only present results with $P_t = 0.25$ in this section as experiments with the other two threshold values did not reveal any substantial differences for tuning the selected hyperparameters of RankNet and RankNEAT.

\newcommand{\respw}{0.3\textwidth}

\begin{figure*}[!tb]
\centering
\subfloat[Endless: $P_t = 0.5$]{\includegraphics[width=\respw]{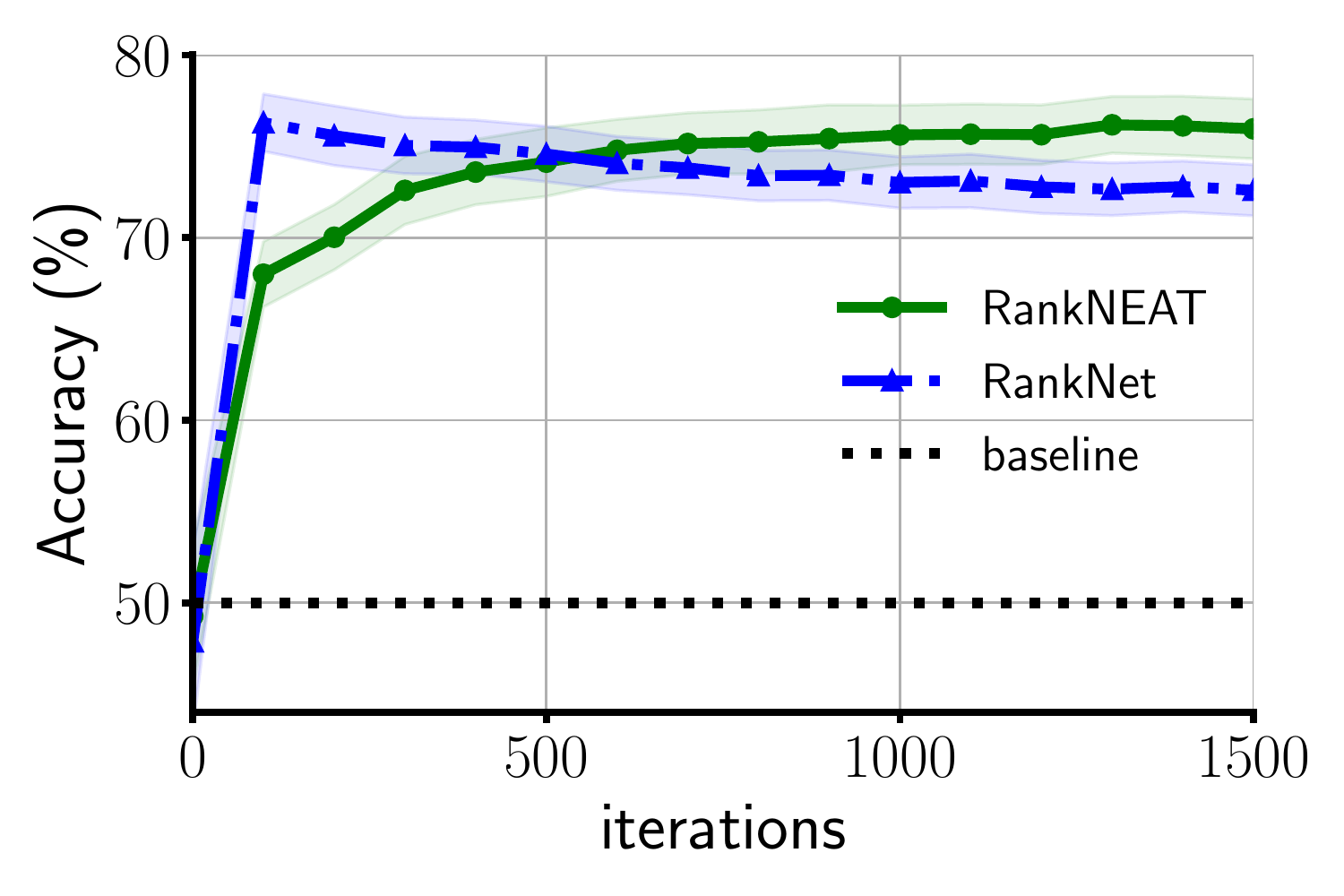}\label{fig:7a}}\quad
\subfloat[Endless: $P_t = 0.25$]{\includegraphics[width=\respw]{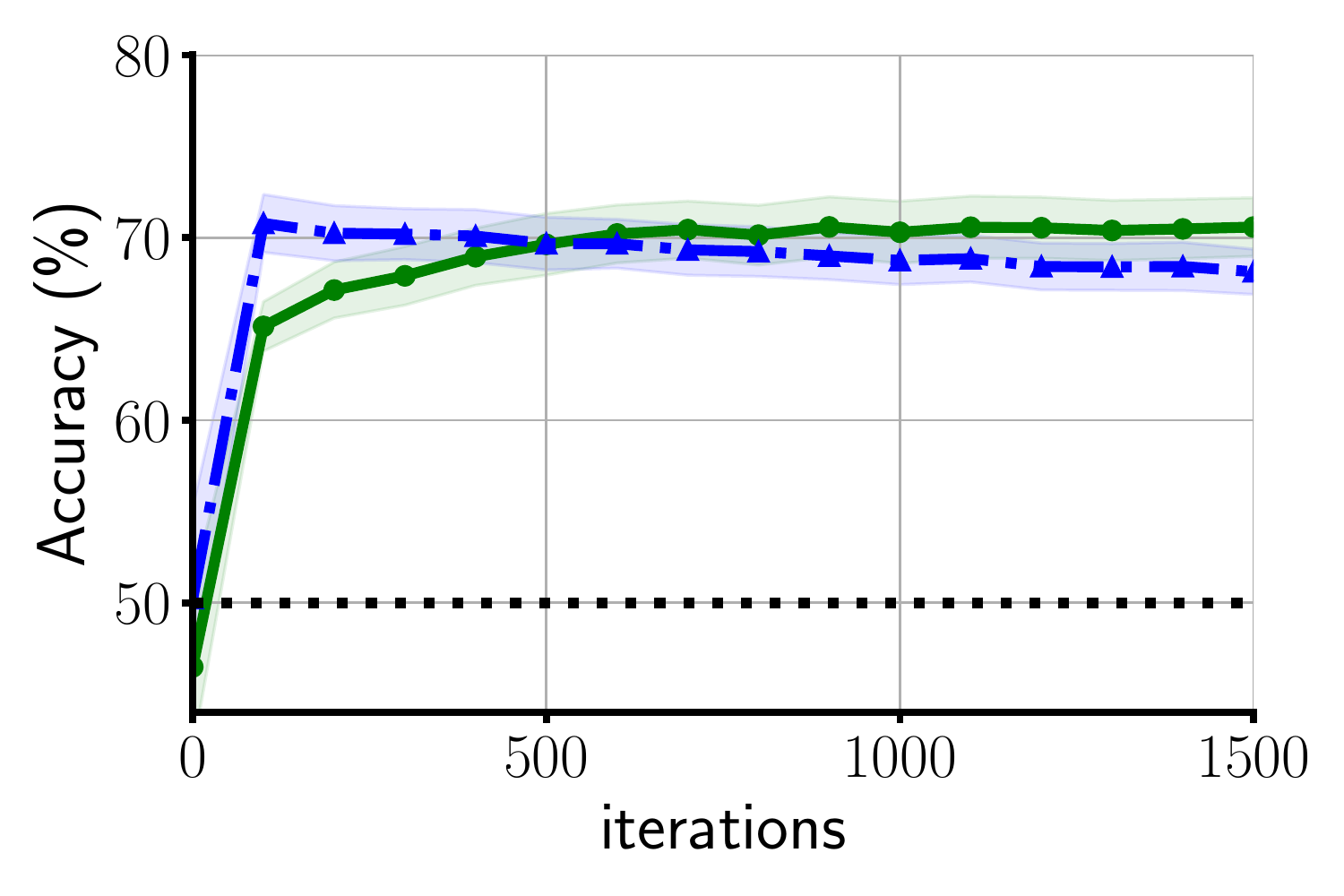}\label{fig:7b}}\quad
\subfloat[Endless: $P_t = 0.15$]{\includegraphics[width=\respw]{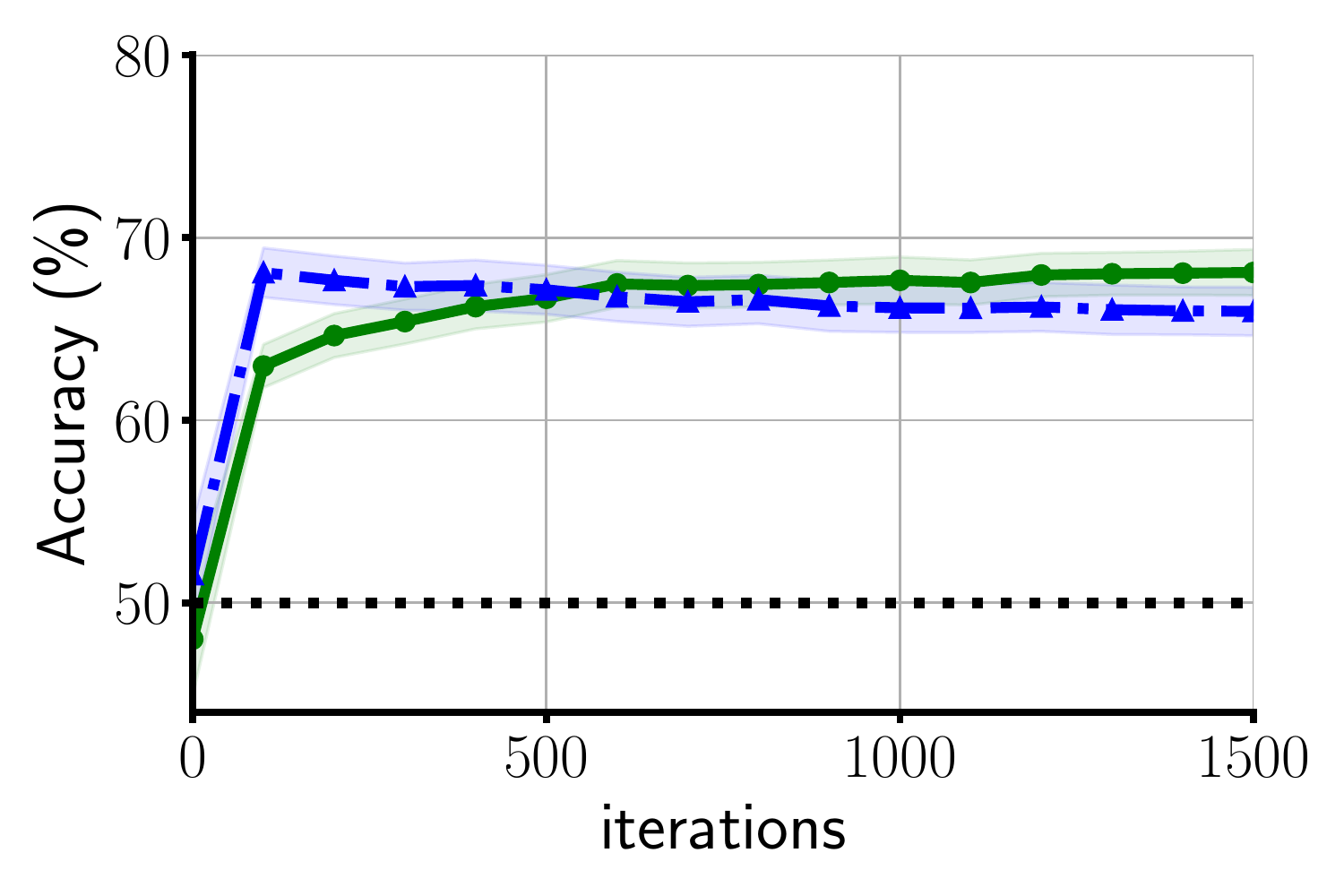}\label{fig:7c}}\\
\subfloat[Pirates: $P_t = 0.5$]{\includegraphics[width=\respw]{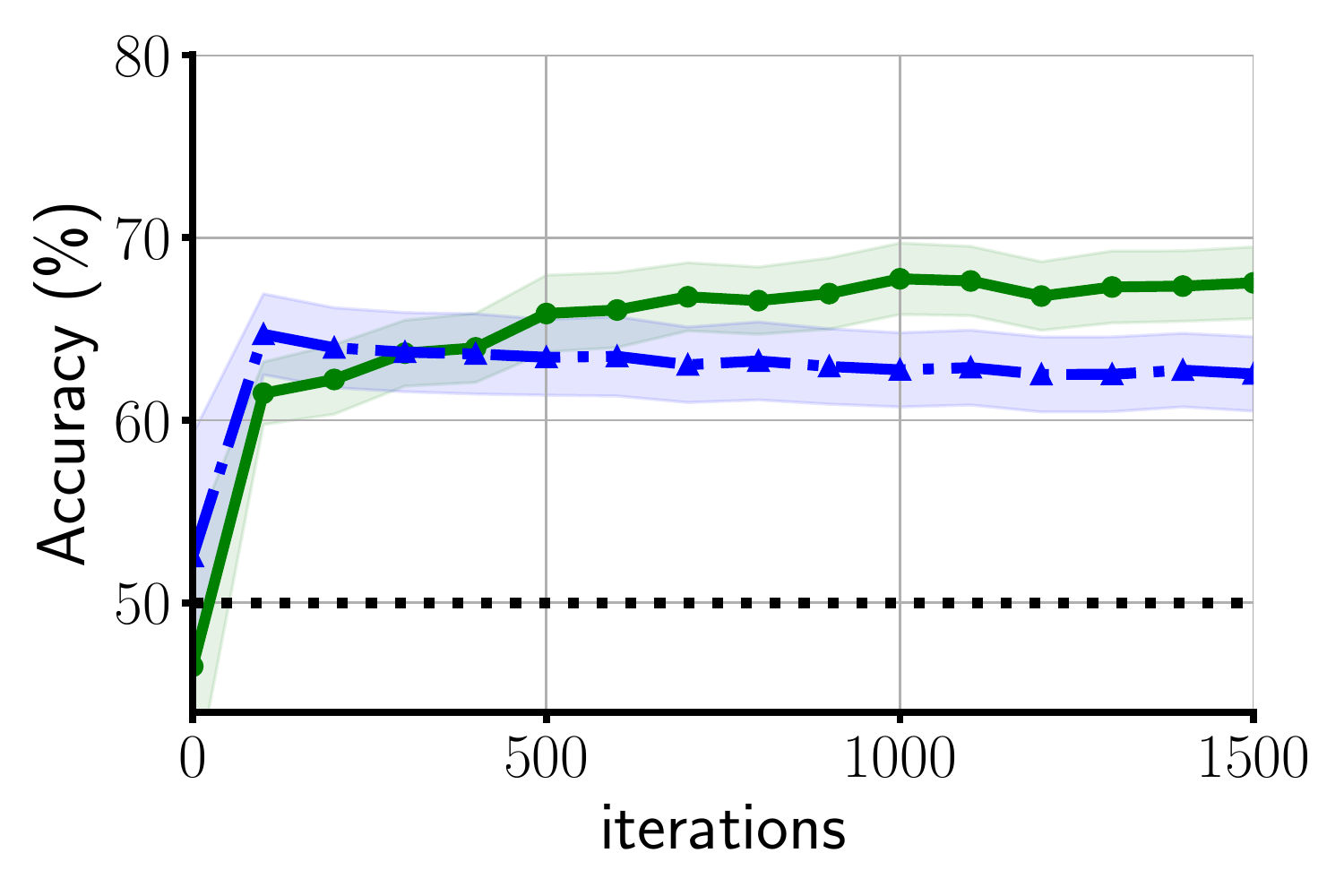}\label{fig:7d}}\quad
\subfloat[Pirates: $P_t = 0.25$]{\includegraphics[width=\respw]{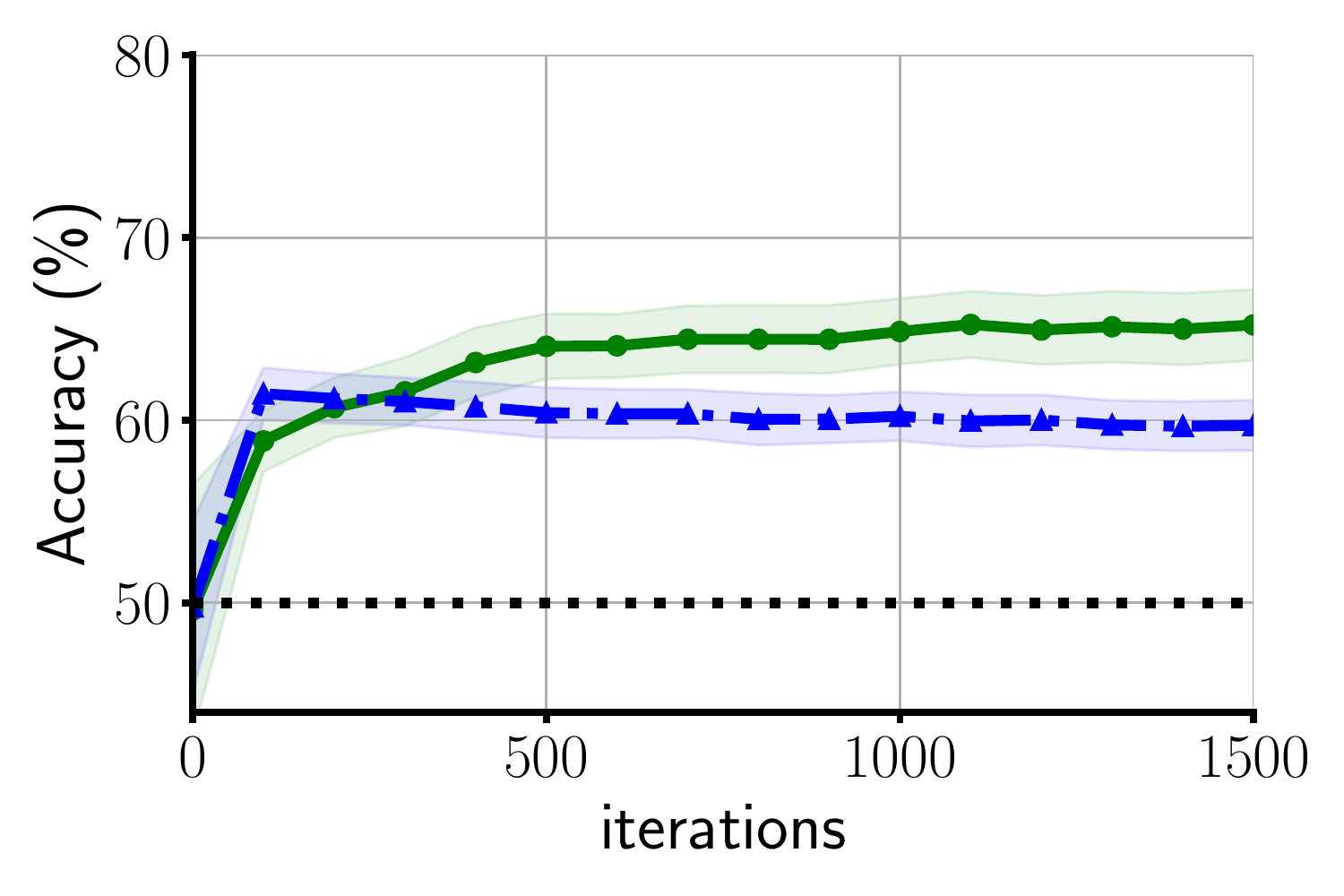}\label{fig:7e}}\quad
\subfloat[Pirates: $P_t = 0.15$]{\includegraphics[width=\respw]{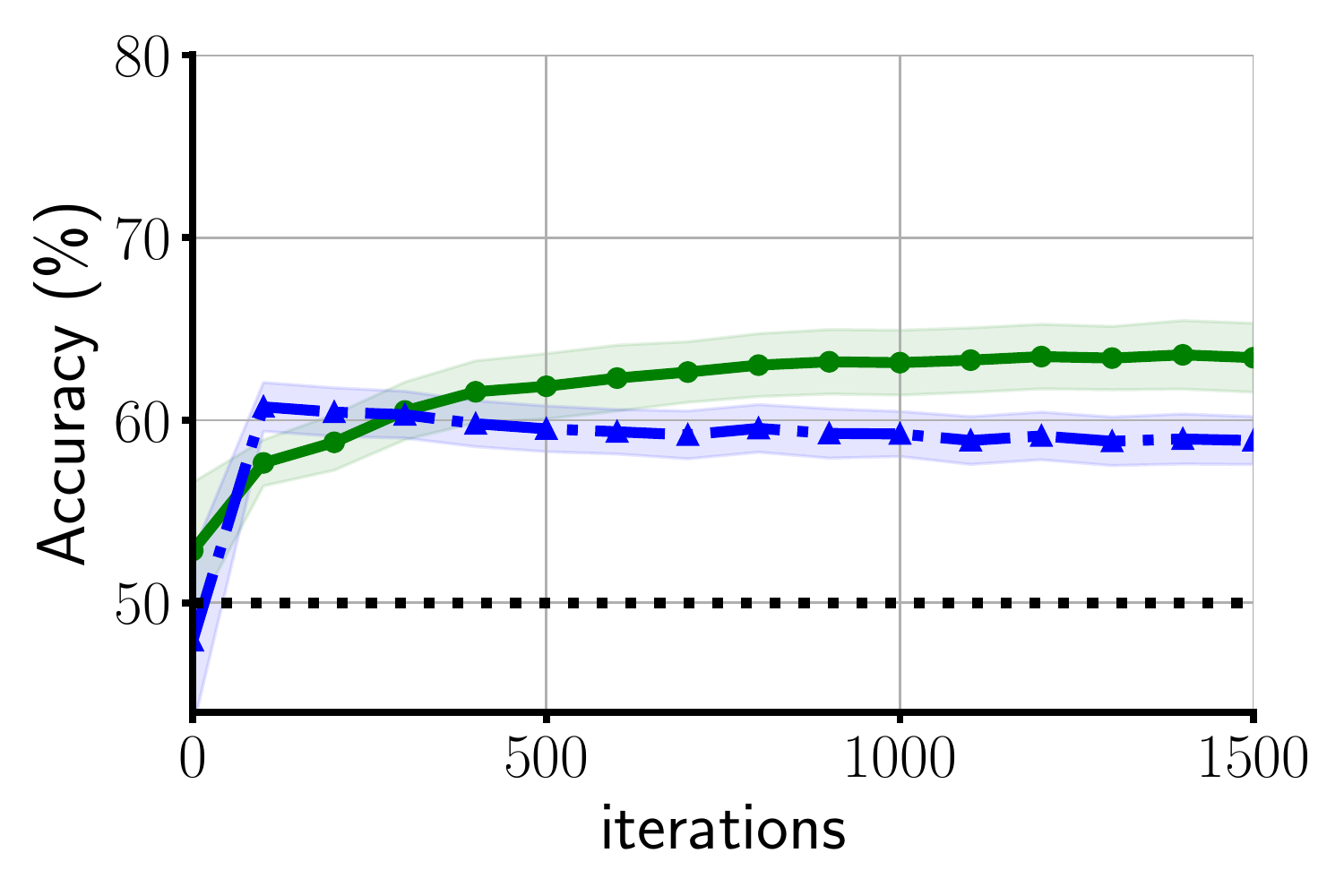}\label{fig:7f}}\\
\subfloat[Run'N'Gun: $P_t = 0.5$]{\includegraphics[width=\respw]{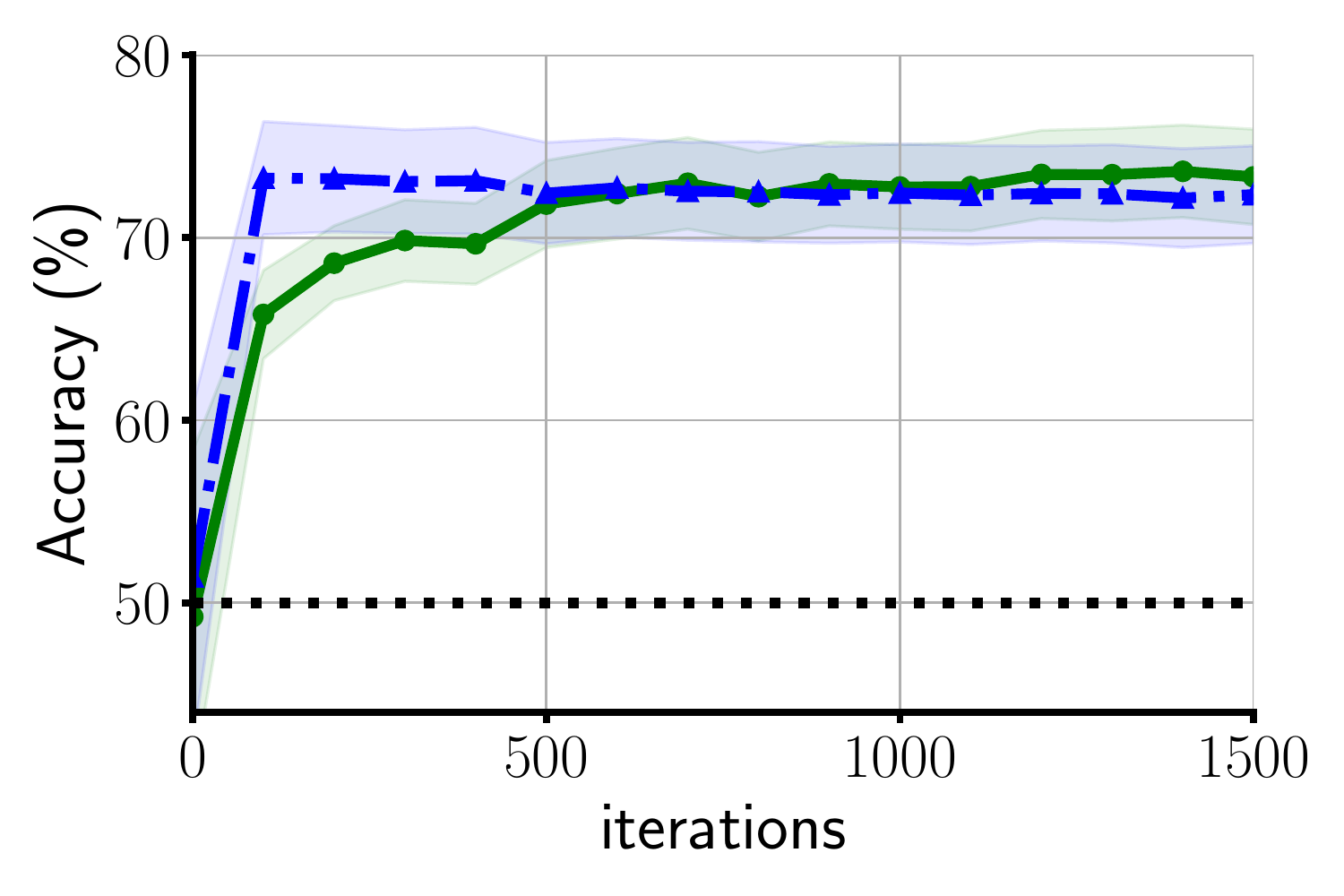}\label{fig:7g}}\quad
\subfloat[Run'N'Gun: $P_t = 0.25$]{\includegraphics[width=\respw]{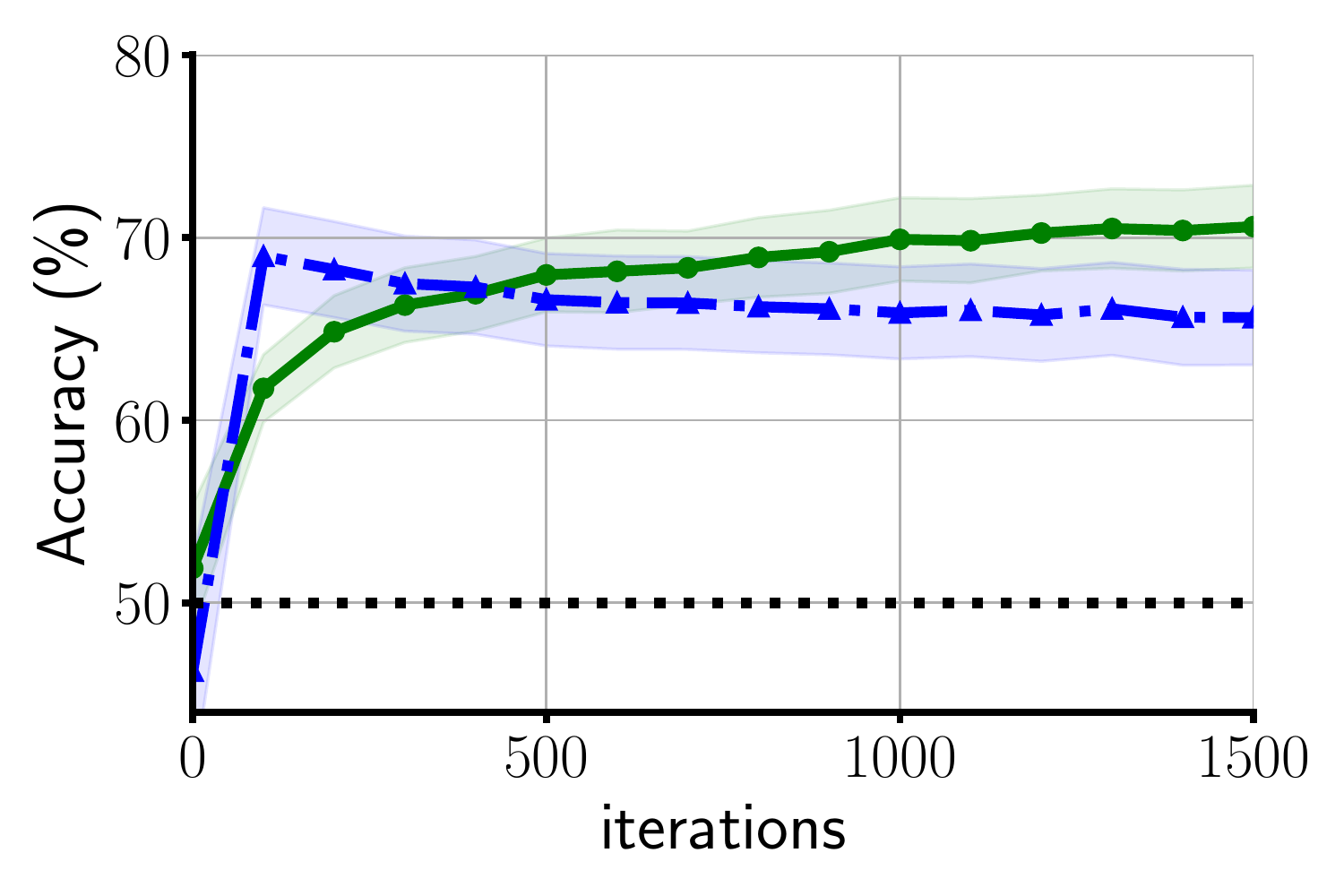}\label{fig:7h}}\quad
\subfloat[Run'N'Gun: $P_t = 0.15$]{\includegraphics[width=\respw]{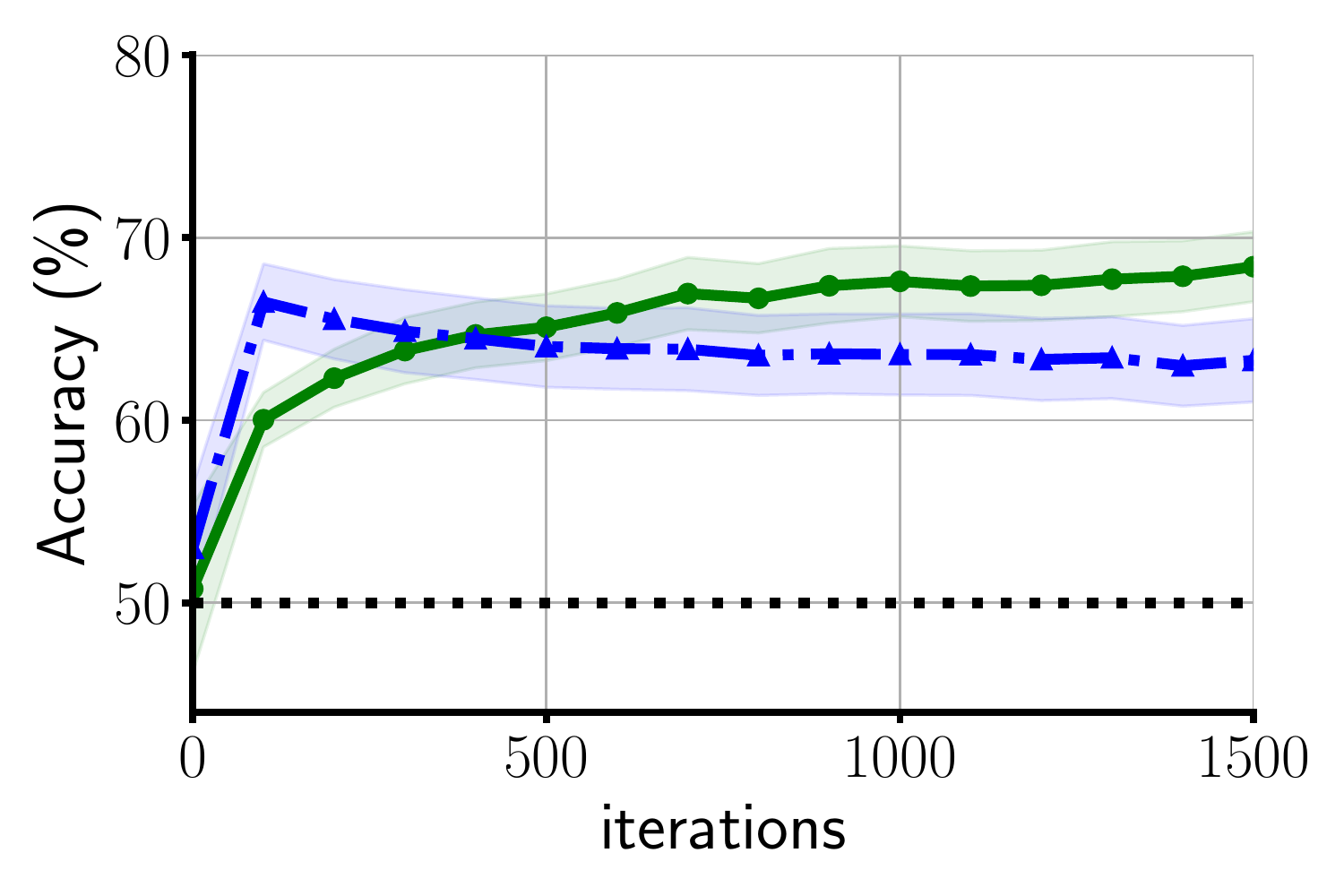}\label{fig:7i}}\\
\caption{Accuracy (and 95\% confidence intervals) over evaluations for the RankNEAT and RankNet models. The black dotted line shows the (random) baseline accuracy of 50\%.}
\label{fig:acc_all}
\end{figure*}

Figure \ref{fig:acc_param_tuning} shows the progress of RankNet (SGD) and RankNEAT (neuroevolution) over 10 epochs and 10 generations, respectively. It should be noted that generations include more evaluations (depending on the population size $p$) than SGD epochs and thus the results between RankNet and RankNEAT are not comparable here. Evidence across all three games shows that large $b_n$ values lead to a quick increase in accuracy for RankNet but subsequent epochs see a drop as the process overfits to the training set. Evidently, with small $b_n$ values testing accuracy increases more slowly but has the potential to reach higher values. Based on this finding, we will use $b_n=10$ as the best parameter in experiments of Section \ref{sec:results_versus}. Evolution on the other hand understandably benefits from larger populations: for instance with $p=1000$ we see a quick optimization at the first generation but relatively small improvements after that. Since with $p=100$ the test accuracy reaches similar values as with $p=1000$ within a few generations, we choose $p=100$ in the experiments reported in the remainder of this paper for its significantly lower computational cost. We should note that optimization for $p=10$ is slow but does not seem to converge within 10 generations, and it is possible that with more generations it could reach the performance of larger populations; however, we could not test this assumption in this paper.

\begin{table*}[!tb]
\caption{Best accuracies (\%) achieved by each model (RankNEAT vs RankNet) for the Endless, Pirates!, and Run'N'Gun test-beds, across three preference threshold values, $P_t$. Values are averaged across 5 independent runs. The average test accuracy of the best run (of 5) is also included within square brackets.}
\label{tab:bestacc}
\centering
\begin{tabular}{|l|c|c|c|c|c|c|}
\cline{2-7} 
\multicolumn{1}{c|}{} & 
\multicolumn{2}{c|}{$P_t=0.5$} & 
\multicolumn{2}{c|}{$P_t=0.25$} & 
\multicolumn{2}{c|}{$P_t=0.15$} \\
\cline{2-7}
\multicolumn{1}{c|}{} & RankNEAT & RankNet & RankNEAT & RankNet & RankNEAT & RankNet \\
\hline
Endless & 76.2 $\pm$1.5	[77.3] &	76.9 $\pm$1.6 [77.9]	&	70.6 $\pm$1.6 [71.7] &	71.5 $\pm$1.6 [72.1]	&	68.1 $\pm$1.3 [69.2]	&	68.5 $\pm$1.3 [69.1]	\\
\hline
Pirates! & 67.8 $\pm$1.9 [69.6]	&	65.8 $\pm$2.2 [66.9]	&	65.2 $\pm$1.8 [66.5] &	62.6 $\pm$1.6 [63.5]	&	63.6 $\pm$1.9 [64.7] &	61.3 $\pm$1.4 [62.1]	\\
\hline
Run'N'Gun &  73.6 $\pm$2.5 [76.3]	&	73.7 $\pm$3.0 [74.8]	&	70.6 $\pm$2.3 [72.3] &	70.2 $\pm$2.3 [71.4] &	68.4 $\pm$1.9 [69.5]	&	67.8 $\pm$2.0 [69.1]	\\
\hline 
\end{tabular}
\label{table:bestacc}
\end{table*}

\subsection{RankNEAT versus RankNet}\label{sec:results_versus}

This section compares the best RankNEAT and RankNet models according to the hyperparameters investigated in Section \ref{sec:results_tuning}. Following earlier comparative studies \cite{mandischer2002comparison} we treat each training epoch and each individuals' fitness evaluation as having the same computational overheads and thus report test accuracy over \emph{iterations} (with each generation of RankNEAT having $p$ iterations, and each epoch in RankNet counting as 1 iteration). As in Section \ref{sec:results_tuning}, we measure test accuracy based on a 10-fold leave-$X$-participants-out cross-validation, repeated and averaged from 5 independent runs. Based on Section \ref{sec:results_tuning}, all RankNet experiments are performed with $b_n=10$ (10 random pairs are sampled from the training set per epoch to calculate the gradient) and all RankNEAT experiments are performed with $p=100$ (100 individuals in the population).

Figure \ref{fig:acc_all} shows the progress over many iterations for the different datasets produced from different games and different preference thresholds ($P_t$). Even though we chose $b_n=10$ because it did not overfit during the short training runs of Section \ref{sec:results_tuning}, it is evident that as training progresses RankNet still is prone to overfitting. In all cases, test accuracy for RankNet drops after the first 100 iterations, often significantly (e.g. in Fig.~\ref{fig:7a}). On the other hand, evolution starts performing poorly but steadily increases at later generations. While evolution assesses its individuals in terms of accuracy in the training set and consistently improves there, it is evident that the models are also able to perform well (despite some fluctuations between generations) in the test set. At the same computational effort ($1,500$ iterations), RankNEAT yields between $1\%$ and $5\%$ higher test accuracies from RankNet, on average, across the 9 experiments performed (with RankNEAT significantly outperforming RankNet in 5 of our 9 tests). Admittedly, in some of the experiments this is due to a noticeable drop in accuracy at later epochs for RankNet; in practice an early stopping criterion for RankNet would likely prevent this. Taking the best models discovered, on average, within these $1,500$ iterations as a whole, we derive the results of Table \ref{table:bestacc}. Here, we see that the results are comparable in several cases, although for the \emph{Pirates!} game RankNEAT consistently performs better. It is worth noting that all models regardless of method underperform in Pirates! We hypothesize that RankNEAT may be able to perform better in more challenging problems. It is also worth noting that when we compare the best run of each algorithm, RankNEAT yields higher accuracies than RankNet in 7 out of 9 experiments.

Apart from the fact that RankNEAT performs global optimization, we expect that the custom operators that add or delete edges are especially powerful for this problem. As noted in Section \ref{sec:preference_learning}, our version of RankNEAT does not allow for larger topologies to emerge but both speciation and topology changes in the edges are expected to have an impact. We expect that deleting an edge can act as a feature elimination mechanism and remove features that do not play a role in predicting arousal. Indeed, we observe that the best models of Table \ref{tab:bestacc} for RankNEAT have between 5\% and 6\% fewer edges than the fully connected SGD network (RankNet with 768 edges). Due to the stochastic nature of the edge removal operator, this ``feature selection'' requires several generations to be impactful, but may largely be responsible for the good performance of the models. 


We observe that models tend to be more accurate at higher preference thresholds. This is not surprising, and matches past findings \cite{makantasis2019pixels}, as ambiguous rankings are more aggressively cleaned. It is worth noting, however, that this comes at the cost of volume and generality of the dataset: indicatively, the datapoints at $P_t=0.50$ are only 28\% of the datapoints at $P_t=0.15$ across all games.

\subsection{Qualitative findings}\label{sec:results_activation}

\newcommand{\actw}{0.22\textwidth}
\begin{figure}[!tb]
\centering
\includegraphics[width=0.95\linewidth]{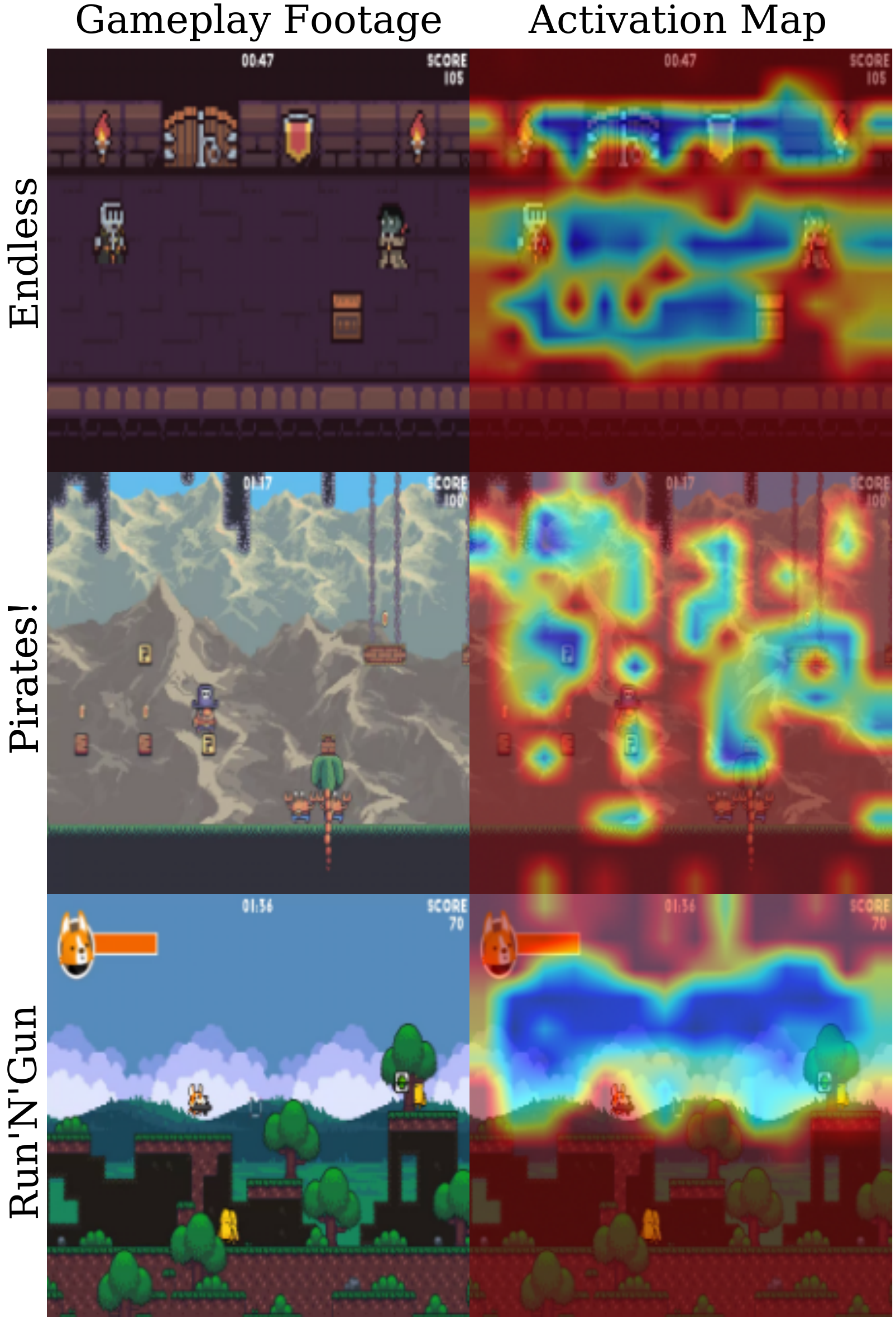}
\caption{Eigen-CAMs for indicative frames of each game: saturated areas show pixels that are important predictors of arousal.}
\label{fig:activation}
\end{figure}

Results presented in the previous section show that player arousal can be modeled based on general-purpose representations such as video frames and, consequently, pixels. Drawing inspiration from the study of Makantasis et al. \cite{makantasis2019pixels}, we constructed the class activation maps (CAM) in order gain insights on which regions of the frames contributed the most to the final result. Our Eigen-CAM implementation relies heavily on the \texttt{PyTorch} library for CAM methods \cite{jacobgilpytorchcam}. It should be noted that Eigen-CAM visualizes the principal components of the learned features, and thus it does not rely on the backpropagation of gradients or any other class relevance score \cite{muhammad2020eigen}. We use RankNEAT, as it achieves the highest accuracies overall, to construct the visualization of Fig.~\ref{fig:activation}. In these activation maps, warmer colors correspond to higher predictors of arousal value for a specific player in the test set. From the samples of Fig.~\ref{fig:activation}, we observe that important predictors of arousal across games are regions containing information about the player, such as the avatar's position, life, game time, and score. Furthermore, the regions that contain information about the enemies' avatars are also very important for the model. In two out of three games, the model manages to mask out some of the redundant information in the environment, such as empty space in Endless or the sky background in Run'N'Gun. For Pirates!, however, such patterns are less clear, and the model precludes the powerups from high importance regions. This may explain the relatively low accuracy value achieved on this game. 


\section{Discussion}\label{sec:discussion}

This work investigated the potential of neuroevolution for handling PL tasks when labels are defined in a subjective and ill-posed manner. We aimed to assess the power of NEAT as a preference learner by comparing the accuracy of NEAT and backpropagation in arousal prediction from general-purpose representations (gameplay videos) across three platformer games. To the best of our knowledge, this is the first time a NEAT algorithm has been used in a PL task. In particular, we studied the case of player affect modeling due to the fact that capturing the emotional manifestation of players is of great import for the domain of digital games \cite{yannakakis2011experience}\cite{yannakakis2005player}. The experiments indicate that RankNEAT can outperform RankNet by avoiding overfitting. There is evidence that RankNEAT's operators for deleting or adding edges is beneficial as a form of feature selection.

It should be noted that there is no straightforward way to compare evolution and SGD methods fairly. While past approaches have used CPU time \cite{mandischer2002comparison,yannakakis2003performance,yannakakis2007emerging}, we instead matched epochs and individual evaluations as approximations of effort. That said, SGD selects a subset of the training data (in experiments in Section \ref{sec:results_versus}, this was $b_n=10$) to derive a gradient while evolution evaluates cross-entropy in all pairings of the training set. Because we could multi-thread the evaluation of individuals in each generation, RankNEAT was between 57\% and 72\% faster in CPU times than RankNet per run, for the $1,500$ iterations of Section \ref{sec:results_versus} (tested on a CPU-only Intel Xeon, 132GB RAM). We could explore different ways of comparing the two methods in future work, as well as perform a more thorough tuning process for the other hyperparameters. In particular, parameters such as the survival threshold, elitism, and minimum species size can affect the crossover stage, increasing the diversity of the population. Thus, properly tuning these parameters may lead to better exploration of the search space. 

Another worthwhile discussion is our choice of applying a more restrained version of NEAT for our experiment. The power of NEAT is arguably the fact that its operators can increase the network size with new nodes and more edges between these new nodes. While our version uses speciation as well as other operators of NEAT, speciation is more meaningful when networks differ in size. The initial population is fully connected while evolved individuals may have fewer edges (i.e. simpler topologies) but never more edges. Preliminary experiments with operators that could add nodes, however, led to an evolutionary process that quickly overfits to the training set while performing poorly on the test set. More experiments are necessary to investigate how this behavior can be countered, e.g. with different fitness evaluation schemes which assess a smaller subset of the training data similar to SGD's batch number. We will also consider and test alternative neuroevolutionary search methods such as covariance matrix adaptation evolution strategy, Differential Evolution, and their respective variants  \cite{neri2010recent,varelas2018comparative} against the introduced algorithm in this paper. When it comes to RankNet there are a plethora of hyperparameters that might influence the performance of a model (e.g. network size, regularization) that need to be examined in a follow up study. The initial study presented here, however, contains a fair amount of hyperparameter tuning experiments for both algorithms as described in our results.

In terms of future research, there are several directions that we can follow to extend the goal of this work. An obvious next step on the scalability of this approach is testing the efficiency of RankNEAT to predict affect for the remaining six games of the AGAIN dataset, which includes racing games and shooter games \cite{melhart2021towards}. A more important next step is testing whether the mapping between pixels and arousal found via neuroevolution can be general-purpose, for instance being able to predict arousal rankings in unseen games of the same genre. Earlier work \cite{melhart2021towards} has shown that gameplay metrics (provided they are well-designed) can be robust predictors of arousal even in unseen games of the same genre. Establishing similar predictors through gameplay footage alone is arguably fundamental for general affect modelling \cite{7860385}. Although this initial study used computer games as its test-bed domain, the proposed method is general and thus applicable to any affective computing and preference learning task; it remains to be found to which degree results hold for other tasks, datasets and domains of preference learning. 

\section{Conclusions}

This paper introduced RankNEAT, an algorithm that transfers the benefits of the NEAT algorithm to learning-to-rank tasks in challenging domains with subjectively defined and biased labels such as affective computing. By leveraging pretrained computer vision models, we were able to evolve accurate models of arousal (with a test accuracy as high as 77\% on average) 
using only gameplay footage. Comparing the performance of neuroevolution against stochastic gradient descent, which is the standard optimization method for PL, we observe that neuroevolution can overcome issues of overfitting. While SGD sometimes can find robust models early on, overfitting leads to a drop in accuracy that is difficult to control for. In contrast, RankNEAT continues to produce ever-more accurate models, and in some cases results had not converged at our ad-hoc cutoff point. Additional experiments, in more games and with more extensive exploration of hyperparameters (such as evolving larger topologies) are necessary to assess the true potential of this approach for player modeling, affective computing, and any machine learning domain that involves human demonstration and annotation.

\begin{acks}
Kosmas Pinitas, Antonios Liapis and Georgios N. Yannakakis were supported by the European Union’s H2020 research and innovation programme (Grant Agreement No. 951911). Konstantinos Makantasis was supported by the European Union’s H2020 research and innovation programme (Grant Agreement No. 101003397). 
\end{acks}

\bibliographystyle{ACM-Reference-Format}
\bibliography{references}

\end{document}